# When Do Typed Decision Models Work in Transportation? Evidence from Travel Choice and Traffic Signal Control

**Hongyang Zhang [a1], Boyuan Deng [b], Siyuan Feng [a, c*2], Kai Chen [d], Junbo Zhang [e, f]**

[a] Department of Aeronautical and Aviation Engineering, The Hong Kong Polytechnic University, Hong Kong, China
[b] Department of Civil and Systems Engineering, Johns Hopkins University, USA
[c] Research Centre for Low Altitude Economy, The Hong Kong Polytechnic University, Hong Kong, China
[d] Department of Computer Science and Engineering, Hong Kong University of Science and Technology, Hong Kong, China
[e] JD iCity, JD Technology, Beijing, China
[f] JD Intelligent Cities Research, China

## Abstract

Typed decision models combine natural-language understanding with a structured interface that returns a selected action and an option-level probability distribution without generating lengthy textual responses. Their low inference cost and response latency make them attractive for transportation problems with predefined alternatives. This paper investigates their effectiveness, robustness, and efficiency in two representative transportation tasks: individual travel-mode choice prediction and sequential traffic signal control. We systematically evaluate TypeSafe Jev and GPT-6 Luna Decisions against conventional transportation methods, reinforcement learning algorithms, and four chat-based large language models. Experiments on the Swissmetro dataset show that typed decision models achieve lower prediction accuracy than traditional discrete choice models and the strongest chat-based baseline, while their performance varies substantially with input representation and question framing. In closed-loop traffic signal control experiments, typed decision models demonstrate competitive performance against established controllers in several scenarios, but exhibit performance degradation under more challenging network conditions. Furthermore, the typed models achieve lower inference latency and computational cost than the evaluated chat-based large language models, highlighting their potential for efficient, zero-shot transportation decision-making. Overall, these findings demonstrate both the potential and limitations of typed decision models, emphasizing the importance of input design, task characteristics, and robustness evaluation for their effective application in transportation systems.

## 1. Introduction

Transportation systems involve a wide range of decision-making problems, from individual travelers selecting among available travel modes to traffic controllers allocating green time among competing traffic movements. Despite their different objectives and temporal characteristics, many of these problems share a common structure: decisions are made from a finite set of predefined alternatives based on contextual information. For instance, travel-mode choice prediction aims to identify the alternative most likely to be selected by a traveler, whereas traffic signal control requires repeatedly selecting feasible signal phases in response to evolving traffic conditions. The former primarily concerns the prediction of human behavior, while the latter involves sequential decisions whose consequences accumulate through system dynamics. Together, these two settings provide complementary perspectives for examining the capabilities of general-purpose decision-making models in transportation.

Traditional transportation decision-making methods are typically developed for specific tasks and operating environments. Discrete choice models, such as multinomial logit (MNL), provide an established framework for modeling individual travel preferences through estimated utility functions **(McFadden, 1974; Train, 2009)**. Traffic signal control has similarly benefited from classical optimization and feedback-based approaches, including fixed-time control and MaxPressure, as well as reinforcement learning (RL) algorithms that learn control policies through interactions with traffic environments **(Varaiya, 2013; Schulman et al., 2017)**. Although these approaches have demonstrated considerable effectiveness, adapting them to new decision contexts may require additional behavioral data, model calibration, simulation, or policy training. More recently, large language models (LLMs) have introduced an alternative paradigm by interpreting natural-language descriptions and selecting actions without necessarily requiring task-specific parameter estimation. Emerging applications in travel behavior modeling and traffic signal control have demonstrated the potential of LLMs as flexible decision-making components **(Nishida et al., 2025; Liu et al., 2026; Lai et al., 2025; Yuan et al., 2025)**. Nevertheless, conventional LLM-based decision-making generally relies on generating and parsing textual responses, which may introduce additional computational overhead and uncertainty in repeated operational decisions.

[1] Email address: hong-yang.zhang@connect.polyu.hk
[2] Corresponding author, email address: siyuan.feng@polyu.edu.hk

A recent development in this direction is the emergence of **typed decision models**, represented by TypeSafe's Jev and OpenAI's GPT-6 Luna Decisions. Unlike conventional chat-based LLMs, these models expose dedicated decision interfaces that accept contextual information, natural-language questions, and explicitly defined candidate alternatives, and return structured answers without generating free-form explanations **(TypeSafe AI, 2026; OpenAI, 2026)**. For choice questions, their outputs include a selected alternative, an option-level probability distribution, and a confidence indicator. Such interfaces offer potential advantages for transportation applications, where feasible actions are frequently predefined and decisions may need to be repeated at short intervals. In particular, they enable decision-making without task-specific parameter training and offer a potentially more efficient alternative to generative LLM inference.

Since the introduction of these models, a growing body of research has begun to examine their decision capabilities and boundaries. **Deußer et al. (2026)** and **Duan et al. (2026)** provide broad evaluations across classification, reasoning, and sequential decision-making tasks, reporting competitive performance in selected settings alongside limitations in long-horizon interactions. Other studies have identified sensitivity to question formulation, probability calibration, and the representation of decision-relevant information **(Meng, 2026; Sharma et al., 2026; Yang et al., 2026)**. These findings suggest that efficient local decision-making does not automatically translate into reliable performance when decisions are embedded in larger workflows or dynamic environments. Nevertheless, existing evaluations have largely focused on general purpose benchmarks and agent-based tasks. How typed decision models perform in transportation systems, where decisions must reflect behavioral preferences or interact with constrained traffic dynamics, remains insufficiently understood.

This study addresses this gap through a systematic benchmark of typed decision models in two representative transportation tasks: travel-mode choice prediction and traffic signal control. The first task evaluates their ability to predict observed human choices using the Swissmetro stated-preference dataset **(Bierlaire et al., 2001; Bierlaire, 2023)**, with particular attention to prediction accuracy, input representation, and question framing. The second task evaluates their ability to make repeated traffic signal decisions in closed-loop SUMO simulations, where individual actions influence subsequent traffic states and system-level performance. We examine two typed decision models, Jev and GPT-6 Luna Decisions, and compare them against traditional transportation methods, reinforcement learning algorithms, and four chat-based LLMs. Beyond decision accuracy and control effectiveness, we investigate the sensitivity of model outputs to input formulations, as well as inference latency and API cost. This evaluation is intended to clarify not only whether typed decision models can produce competitive transportation decisions, but also under what task conditions and input configurations their advantages can be realized.

The contributions of this study are threefold. First, we systematically evaluate the effectiveness and operational efficiency of typed decision models in transportation choice prediction and closed-loop traffic signal control, benchmarking them against traditional transportation methods and chat LLMs to demonstrate their potential for training-free decision-making. Second, we reveal the critical role of input representation in decision quality, showing that alternative representations of the same problem can lead to substantial differences in predictive accuracy and control performance, even without changes to model parameters. Third, we identify a question framing-dependent pattern of option-order sensitivity, demonstrating that the performance varies considerably across input formulations and can be mitigated through appropriate representation design. Together, these findings provide empirical evidence for the capabilities of typed decision models and offer practical insights into their reliable application in transportation systems.

The remainder of this paper reviews related research in Section 2, presents the decision framework in Section 3, describes the experiments in Section 4, and reports results in Sections 5. Section 6 concludes and discusses future work.

## 2. Related work

## 2.1. Travel-mode choice modeling and LLM-based behavioral prediction

Travel-mode choice prediction is fundamental to transportation planning because the selection of private or public transport modes affects demand forecasts, infrastructure evaluation, and policy appraisal. Discrete choice models, especially MNL and its random-utility extensions, estimate how travel time, cost, service frequency, and traveler characteristics affect individual choices **(McFadden, 1974; Train, 2009)**. Their parameters are inferred from labeled choice observations. This produces interpretable estimates and strong prediction performance in settings where relevant data and a suitable specification are available, but the estimation process must be repeated or adapted when alternatives and traveler populations change.

Travel-mode choice modeling plays an important role in transportation demand forecasting and policy evaluation. Rooted in random utility theory, discrete choice models represent individual decisions as the outcome of comparing the utilities of available alternatives. The multinomial logit (MNL) model and its extensions have been widely adopted to estimate the effects of travel time, monetary cost, service attributes, and socioeconomic characteristics on travelers' preferences **(McFadden, 1974; Train, 2009)**. Their main strengths include interpretable behavioral parameters and a well-established statistical foundation. However, the performance depends on the specification of utility functions and the availability of representative choice observations. Machine-learning methods have subsequently been introduced to capture nonlinear relationships and heterogeneous preferences. **Hagenauer and Helbich (2017)** investigated machine-learning classifiers for travel-mode choice, while **Salas et al. (2022)** compared machine-learning approaches and discrete choice models under behavioral heterogeneity. These studies demonstrate the potential of data-driven prediction while highlighting the importance of model specification and training data.

Recent studies have explored LLMs as general-purpose predictors of human travel behavior. **Nishida et al. (2025)** investigated the ability of LLMs to predict transportation mode choices across different sets of alternatives, seeking greater flexibility than models estimated for a fixed choice set. **Liu et al. (2026)** proposed a persona-based approach to align LLM-generated travel choices with observed human behavior, using traveler characteristics to improve behavioral consistency. **Mo et al. (2026)** systematically compared zero-shot prompting, prompt optimization, and embedding-based prediction against conventional supervised methods across multiple travel-behavior datasets. Their findings suggest that LLM-based approaches can be particularly attractive when labeled data are limited. Beyond mode choice, **Wang et al. (2025)** developed an LLM-based agent framework for day-to-day route choice, illustrating the broader potential of language models in behavioral transportation simulation. These researches offer a potential route to decision-making with little or no task-specific fitting, but chat-based decision extraction still requires text generation and parsing.

To summary, previous studies have explored LLMs for travel-mode choice, but the performance of typed decision models remains unclear. In particular, can these models predict observed choices without task-specific training while lowering inference cost and latency, and how sensitive are their predictions to different input descriptions? We examine these questions using the public state of preference dataset.

### 2.2. Traffic signal control: adaptability versus training and inference overhead

Traffic signal control is a sequential decision-making problem in which local phase selections influence vehicle queues, downstream congestion, and subsequent traffic conditions. Classical methods determine signal timing through predefined plans or adaptive rules. **Webster (1958)** developed a foundational method for fixed-time signal timing. Pressure-based control provides an adaptive alternative by selecting signal phases based on upstream and downstream traffic conditions. **Wongpiromsarn et al. (2012)** proposed a distributed backpressure controller with network throughput guarantees, while **Varaiya (2013)** established the MaxPressure framework for signalized networks. Subsequent studies extended pressure-based control to account for finite link capacities and congestion spillback **(Gregoire et al., 2015)**, cyclic signal operation **(Le et al., 2015)**, and coordination along urban corridors **(Xu et al., 2024)**. These studies demonstrate the effectiveness of pressure-based control without requiring task-specific policy training.

Reinforcement learning (RL) has been widely applied to traffic signal control by learning phase-selection policies through interactions with traffic environments. IntelliLight used deep Q-learning to optimize traffic signals based on observed traffic conditions **(Wei et al., 2018)**. FRAP introduced phase competition modeling to improve learning efficiency and generalization across intersection layouts **(Zheng et al., 2019)**. PressLight incorporated MaxPressure principles into RL state and reward design **(Wei et al., 2019)**. For network-level control, CoLight used graph attention to coordinate neighboring intersections **(Wei et al., 2019)**, while MPLight extended pressure-based RL to large-scale traffic networks **(Chen et al., 2020)**. Subsequent studies addressed adaptability and scalability. AttendLight developed an attention-based policy for different intersection configurations **(Oroojlooy et al., 2020)**, and GPLight introduced grouped multi-agent learning to balance coordination and computational complexity **(Liu et al., 2023)**. PPO-based controllers have also been studied for adaptive signal control **(Schulman et al., 2017; Huang and Qu, 2023)**. These methods demonstrate the effectiveness of RL but generally require simulation-based training and careful validation across traffic conditions.

More recently, LLMs have been explored for traffic signal control. **Movahedi and Choi (2025)** developed LLM-based agents using zero-shot reasoning and accumulated traffic knowledge for adaptive control. LLMLight used traffic-aware prompting and introduced LightGPT, a specialized language model for signal decisions **(Lai et al., 2025)**. CoLLMLight extended this approach to network-wide control through cooperation among LLM agents **(Yuan et al., 2026)**. Other studies have combined LLMs with RL. iLLM-TSC used an LLM to evaluate and correct RL decisions under imperfect observations and unexpected traffic events **(Pang et al., 2026)**, while VLMLight introduced an LLM-based meta-controller that switches between RL and safety-oriented reasoning **(Wang et al., 2025)**. More recently, Traffic-R1 used reinforcement learning to improve language-model reasoning and multi-intersection coordination **(Zou et al., 2026)**. ProcessLight further introduced step-level supervision to improve the reasoning process for traffic signal decisions **(Zhao et al., 2026)**. These studies demonstrate the potential of LLMs for traffic control, but also highlight the importance of specialized training, coordination, inference efficiency, and reliable decision-making.

Typed decision models offer a different implementation of language-based control: instead of generating an action through an open-ended textual response, they directly select from a predefined set of feasible signal phases. This structure is attractive for traffic signal control because the action space is naturally discrete. The quality of each decision depends on the available traffic observations, while overall network performance depends on the accumulated consequences of repeated phase selections. Existing work on LLM-based traffic control has emphasized specialized reasoning, training, or inter-intersection cooperation. In contrast, this study evaluates general-purpose typed decision models directly, without task-specific training or explicit coordination mechanisms, to establish their baseline capabilities and limitations under closed-loop traffic simulation.

## 2.3. General-purpose typed decision models and their reliability

Typed decision models have recently emerged as an alternative to conventional text-generating LLMs for tasks that require predefined output types. Rather than producing an unrestricted natural-language response, a model receives contextual evidence and a structured question and returns an answer within a specified decision space. TypeSafe's Jev and OpenAI's Decisions API support this general approach, including fixed-choice questions accompanied by option-level probability distributions and confidence indicators **(TypeSafe AI, 2026; OpenAI, 2026)**. Unlike instructing a generative LLM to return a JSON-formatted response, these systems implement structured decisions as part of their inference interfaces. Their potential advantages include reduced output-generation overhead and direct integration with downstream software that requires discrete, machine-readable decisions. Whether these advantages translate into improved decision quality remains an empirical question.

A growing number of studies have examined the general-purpose decision capabilities of these models. **Deußer et al. (2026)** evaluated Jev across 37 datasets spanning classification, language understanding, reasoning, and related tasks, demonstrating competitive performance and low inference cost in several settings. Their extended evaluation also considered sequential decision environments, revealing limitations when individually selected actions must be composed into longer trajectories. **Duan et al. (2026)** introduced JEVal, a benchmark covering 36 datasets across multiple application domains, and compared general decision models with generative LLMs. Their results indicated that decision models can be effective when the required judgment is supported by the available evidence, but may experience reliability problems in long-horizon interactions. In another application, **Deng et al. (2026)** examined Jev as a semantic decision component in scientific workflows, distinguishing the correctness of intermediate choices from their downstream computational consequences. Together, these studies provide evidence that typed decision models can be efficient components of structured decision workflows, while emphasizing the need to evaluate more than the correctness of isolated answers.

Beyond overall accuracy, input sensitivity and probability reliability have emerged as important research issues. Previous work on generative LLMs established that seemingly minor changes in prompt formatting and candidate order can alter model responses **(Sclar et al., 2024; Pezeshkpour and Hruschka, 2024)**. Recent evaluations suggest that similar concerns remain relevant for typed decision interfaces. **Meng (2026)** audited the probability outputs of Jev and found that calibration behavior can vary across tasks and question formulations. **Sharma et al. (2026)** introduced TypedBench to assess decision accuracy, framing sensitivity, probability quality, and downstream decision costs, demonstrating that performance and probability reliability cannot be characterized by accuracy alone. **Munot (2026)** further investigated probability allocation in Jev, Liquid d1, and GPT-6 Luna Decisions, showing that the distribution of probability mass across multiple valid alternatives can affect threshold-based decision rules. These findings motivate the evaluation of typed decision models under alternative input representations and question formulations, rather than relying on a single prompt configuration.

Another emerging research direction concerns the use of typed decision models in sequential tasks. **Yang et al. (2026)** investigated the limitations of one-pass decision models and showed that performance can deteriorate when a decision requires reconstructing information through multiple updates or when the supplied state summary is unreliable. **Deußer et al. (2026)** and **Duan et al. (2026)** also reported limitations in interactive environments, where locally reasonable decisions do not necessarily lead to successful task completion. More recently, **H. Liu et al. (2026)** examined decision models as actors, judges, and teachers in long-horizon agent tasks, showing that their ability to evaluate actions or provide learning signals may differ from their effectiveness as direct action-selection policies. These findings reveal an important distinction between evaluating a single decision and evaluating the outcome of a sequence of interdependent decisions. For transportation applications, this distinction is particularly relevant because each control action changes traffic conditions and influences the consequences of subsequent actions.

In summary, existing studies have established a solid foundation for travel choice prediction and traffic signal control through traditional statistical models, optimization methods, and learning-based methods. More recently, large language models have demonstrated their potential for transportation decision-making, although their performance and practical deployment remain subject to challenges related to inference efficiency, input sensitivity, and decision reliability. Typed decision models offer a promising alternative by combining natural-language understanding with structured decision outputs and efficient inference. Despite these advances, existing typed decision benchmarks provide limited evidence regarding their effectiveness in transportation systems. Travel behavior prediction requires agreement with heterogeneous human preferences rather than merely selecting an objectively preferable alternative. Traffic signal control requires repeated choices under physical constraints, interacting complicated and dynamic environment, and delayed network-wide consequences. To address these gaps, this study conducts a systematic evaluation of typed decision models across travel mode choice prediction and traffic signal control, examining their effectiveness, robustness, and computational efficiency under different decision settings.

## 3. Problem formulation and evaluation framework

### 3.1. Typed choice as a common decision interface

Consider a decision instance with state $s_t$ and a finite set of allowable actions $\mathcal{A}_t = \{a_1, \dots, a_J\}$. The application transforms the original information into an input description $x_t = \phi(s_t)$ and asks a natural-language question $q_t$. A typed decision model returns a selected action, an option-level distribution, and a confidence score:

$$(\hat{a}_t, \tilde{\mathbf{p}}_t, \kappa_t) = f_\theta(x_t, q_t, \mathcal{A}_t), \qquad \hat{a}_t \in \mathcal{A}_t, \quad \tilde{\mathbf{p}}_t \in [0,1]^J.$$

The selected option is executed directly or used as the prediction, depending on the application. The distribution $\tilde{\mathbf{p}}_t$ contains the API's reported option probabilities and $\kappa_t$ its confidence indicator. A chat LLM solves the same finite-choice problem by producing a textual or JSON-formatted option label, which the application parses. Conventional methods instead calculate the choice through an estimated utility function, a predefined rule, or a learned policy.

A mode-choice request has the same concrete form. Suppose the state describes a business traveler aged 40–54, with income under CHF 50k, traveling without luggage and without a GA annual pass. The candidate set is {TRAIN, SWISSMETRO, CAR}. TRAIN is described as 213 min and CHF 100, every 30 min; SWISSMETRO as 109 min and CHF 114, every 10 min; and CAR as 144 min and CHF 130, with the relative comparisons explicit: TRAIN is 104 min slower and CHF 14 cheaper than SWISSMETRO, while SWISSMETRO is 35 min faster and CHF 16 cheaper than CAR. The question is "Which alternative did this traveler choose in this stated-preference survey? Choose the alternative that best matches the traveler's behavior, not the one you would recommend." A schematic response would be {"choice": "SWISSMETRO", "probabilities": {"TRAIN": 0.14, "SWISSMETRO": 0.68, "CAR":

0.18}, "confidence": 0.52}. This is an interface illustration, not a recorded model output; Appendix A reproduces an exact, code-generated request/response pair from the experiment.

A signal-control request has a similar structure. Suppose the state at one 2-way epoch reports that phase 0 is currently active, phase 1 serves 12 queued vehicles with a 35 s longest wait, phase 2 serves 5 vehicles with a 70 s longest wait, and phase 3 serves 2 vehicles with a 110 s longest wait; the candidate set is {phase_0, phase_1, phase_2, phase_3}, and the question asks which phase should receive green next. A schematic response would be {"choice": "phase_3", "probabilities": {"phase_0": 0.08, "phase_1": 0.17, "phase_2": 0.21, "phase_3": 0.54}, "confidence": 0.41}. This is also a schematic illustration, not a recorded output. The controller uses the server-selected choice, maps phase_i to action i, and applies the shared max-green wrapper; Appendix A contains the exact request/response pair from the same 2-way decision epoch.

### 3.2. Static observed-choice prediction

Let $y_n \in \mathcal{M}_n$ denote the observed travel mode of choice observation $n$, where $\mathcal{M}_n$ is its available mode set, and let $\hat{y}_n$ be the predicted mode. We measure individual prediction accuracy by

$$\mathrm{Top1} = \frac{1}{N}\sum_{n=1}^{N} \mathbf{1}\{\hat{y}_n = y_n\}$$

For each mode $m$, precision $P_m$ is the proportion of predicted-$m$ observations actually belonging to $m$, and recall $R_m$ is the proportion of observed-$m$ choices correctly identified. Macro-F1 gives equal weight to the modes:

$$F1_m = \frac{2P_m R_m}{P_m + R_m}, \qquad \mathrm{MacroF1} = \frac{1}{|\mathcal{M}|}\sum_{m\in\mathcal{M}} F1_m$$

We also report choice-share mean absolute error (Share MAE), which compares the predicted and observed fractions of choice occasions assigned to each alternative:

$$\hat{s}_m = \frac{1}{N}\sum_{n=1}^{N} \mathbf{1}\{\hat{y}_n = m\}, \quad s_m = \frac{1}{N}\sum_{n=1}^{N} \mathbf{1}\{y_n = m\}, \quad \mathrm{ShareMAE} = \frac{1}{|\mathcal{M}|}\sum_{m\in\mathcal{M}} |\hat{s}_m - s_m|$$

A lower Share MAE means that the model better reproduces the overall mode shares, although it may still misclassify individual travelers. Macro-F1 and Share MAE complement Top-1 because the three choices are not equally frequent in Swissmetro.

We employ a multinomial logit (MNL) model as a conventional discrete choice benchmark for travel mode choice prediction. Following the random utility framework, the systematic utility of alternative (m) for choice observation (n) is specified as

$$V_{nm} = \alpha_m + \beta_T \frac{T_{nm}}{100} + \beta_C \frac{C_{nm}}{100} + \beta_H \frac{H_{nm}}{100}$$

where $\alpha_m$ denotes the alternative-specific constant, and $T_{nm}$, $C_{nm}$, and $H_{nm}$ represent travel time (minutes), travel cost (CHF), and service headway (minutes), respectively. TRAIN is selected as the reference alternative, with $\alpha_{\mathrm{TRAIN}} = 0$, while separate constants are estimated for SWISSMETRO and CAR. As headway is only applicable to TRAIN and SWISSMETRO, we set $H_{n,\mathrm{CAR}} = 0$. All continuous attributes are scaled by 100 for numerical estimation, and the coefficients $\beta_T$, $\beta_C$, and $\beta_H$ are constrained to be non-positive, reflecting the expected disutility of longer travel times, higher costs, and longer service headways. The probability of selecting alternative $m$ is calculated using a softmax function over the available choice set:

$$\Pr(y_n, = m \mid X_n, \mathcal{A}_n) = \frac{\exp(V_{nm})}{\sum_{j\in\mathcal{A}_n} \exp\left(V_{nj}\right)} \mathbf{1}[m \in \mathcal{A}_n]$$

where $\mathcal{A}_n$ denotes the set of available alternatives for observation $n$, ensuring that unavailable alternatives receive zero probability. The model assumes generic coefficients across travel modes without income interactions, mode-specific time or cost sensitivities, or random preference heterogeneity. Parameters are estimated exclusively on the training set and subsequently evaluated on the held-out test set.

### 3.3. Sequential traffic signal control

In traffic signal control, the network state $s_k$ is observed every decision interval $k$. At intersection $i \in \mathcal{I}$, the controller chooses a feasible phase $a_k^i \in \mathcal{A}_i$. The joint action $\mathbf{a}_k = \left(a_k^i\right)_{i\in\mathcal{I}}$ is applied to the simulator $F$ under shared signal constraints, and traffic evolves as

$$s_{k+1} = F(s_k, \mathbf{a}_k, \xi_k)$$

where $\xi_k$ represents exogenous traffic demand and stochastic conditions. A good signal action must therefore be evaluated in combination with subsequent network evolution rather than against a single observed correct label. The main recorded control statistic is the time average of the environment's system-wide mean vehicle waiting-time field. Using $w_{v,k}$ for the simulator-reported waiting-time quantity of vehicle $v$ at step $k$, and $\mathcal{V}_k$ for the vehicles included in that statistic, we write its conceptual aggregation as

$$\overline{W} = \frac{1}{K}\sum_{k=1}^{K}\left(\frac{1}{|\mathcal{V}_k|}\sum_{v\in\mathcal{V}_k} w_{v,k}\right)$$

The corresponding throughput is the number of completed vehicle arrivals per hour of simulated time:

$$Q = \frac{3600 N_{\text{arrived}}}{T}$$

where $N_{\text{arrived}}$ is the simulator's counted arrivals during an episode of $T$ seconds. Lower $\overline{W}$ and higher $Q$ indicate better performance under their respective definitions. We consider both in the control experiments.

## 4. Experimental design

### 4.1. Swissmetro data, baseline models and input variants

The Swissmetro stated-preference survey describes intercity travel situations with three alternatives: TRAIN, SWISSMETRO and CAR. Each choice observation contains traveler-related information and alternative attributes such as travel time, monetary cost, and headway. We use observations with GA = 0, indicating respondents without a Swiss annual public-transport travelcard, and the test sample contains 1,791 choice records from 199 respondents: 144 TRAIN, 1,049 SWISSMETRO and 598 CAR observations. The existing experiment separates respondents into training, validation and test partitions in a reported 60/20/20 ratio to prevent a respondent's observations from appearing across partitions.

The conventional references are MNL, fitted on the training data, and the majority-class rule, which always predicts SWISSMETRO. The evaluated chat LLMs are DeepSeek V4.1 Flash **(DeepSeek, 2026)**, GLM 5.3 Flash **(Z.ai, 2026)**, GPT-6 Luna **(OpenAI, 2026)** and Claude Haiku 5.5 **(Anthropic, 2026)**. These models are deliberately selected from the fast/low-cost tier rather than the strongest available models. The study asks whether typed decision models are viable under deployment-relevant cost and latency; frontier models such as GPT-6 Sol, Claude Opus 5.5, or GLM 5.3 Pro operate at a different cost–latency point and would confound interface effects with model scale. The typed models are likewise cheap and fast: TypeSafe Jev 1.14 and GPT-6 Luna Decisions. This yields a tier-matched comparison against the chat baselines. Both chat LLMs and typed decision models are evaluated directly without fine-tuning or task-specific parameter training.

We examine three aspects of the Swissmetro input: **attribute representation (R)**, **question framing (F)**, and **candidate order**. The five attribute representations are described in Table 1 using a single illustrative choice situation. R0 presents raw descriptions; R1 names attributes in structured fields; R2 adds explicit relative comparisons; R3 replaces magnitudes with coarse categories; R4 uses a narrative sentence. The illustration uses a TRAIN alternative of 213 minutes and CHF 100 and a SWISSMETRO alternative of 109 minutes and CHF 114 as an example.

**Table 1. Examples of input representations for a TRAIN alternative (illustrative).**

| Representation | Example of the supplied option description |
| --- | --- |
| R0 — Raw | TRAIN: 213 min; CHF 100; headway 30 min |
| R1 — Structured | {"travel_time_min":213,"trip_cost_chf":100,"headway_min":30} |
| R2 — Relative | TRAIN is 104 min slower and CHF 14 cheaper than SWISSMETRO. |

| Representation | Example of the supplied option description |
|---|---|
| R3 — Bucket | TRAIN: long travel time (>120 min); expensive (>CHF 60); service every 30 min. |
| R4 — Narrative | The traveler takes a 213-minute train journey costing CHF 100, with departures every 30 minutes. |

For R2, three framings were tested. F1 asks, *'Which alternative did this traveler choose?'*, targeting the observed survey response. F2 asks, *'Which alternative would this traveler choose?'*, requesting a hypothetical prediction. F3 asks, *'Which alternative is best for this traveler?'*, asking for a recommendation. Finally, the candidate/option order is kept original, reversed, or rotated while all descriptions are preserved. The main observed-choice comparison uses F1. The framing and order tests are reported separately.

### 4.2. SUMO signal-control environments

We use the microscopic traffic simulator SUMO **(Lopez et al., 2018)**, with the SUMO-RL environment interface **(Alegre, 2019)**, for four network scenarios. 2way contains one four-way intersection with four green phases. cologne1 is a map-derived single-intersection network with four phases. 2x2 comprises four intersections arranged as a grid, with four phases each. cologne3 contains three intersections with 3, 4 and 4 admissible phases. The latter two scenarios test multi-intersection signal decisions under interdependent traffic conditions.

Each run lasts 1,800 simulation seconds with a nominal decision interval of 5 seconds, yielding 360 control epochs. The experiment uses three traffic seeds (42, 43, and 44) and a signal wrapper that sets minimum green to 5 s, maximum green to 50 s, and yellow transitions to 2 s. The recorded route-demand summaries are approximately 202, 219, 140 and 332 vehicles for 2way, cologne1, 2x2 and cologne3, respectively. For the multi-intersection cases, each controller first selects one phase for each intersection; the joint action is then executed together in the simulation step.

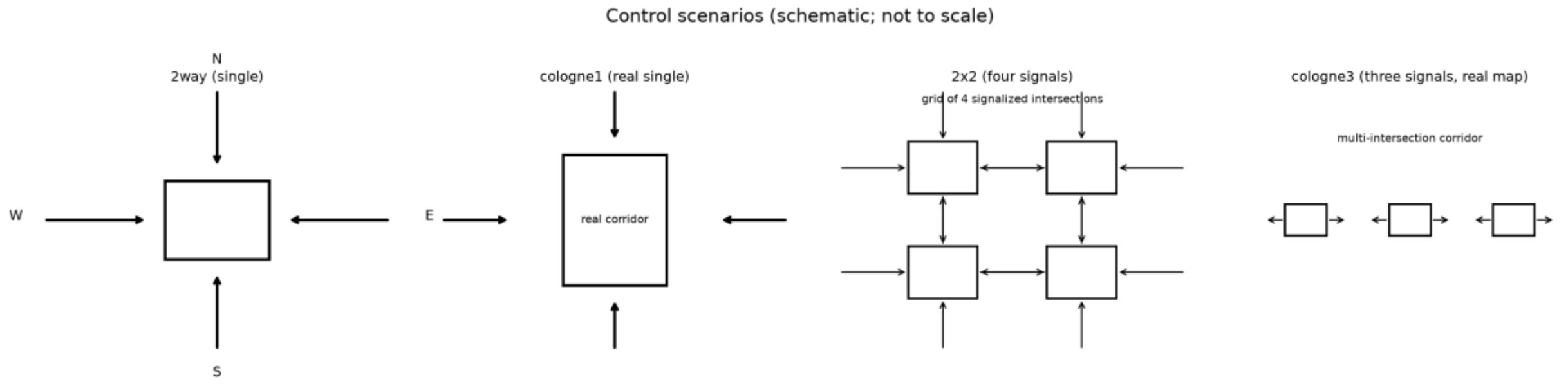

**Figure 1.** Control scenarios schematic

### 4.3. Compared traffic signal controllers

We evaluate four categories of controllers: traditional traffic-control rules, RL-based control, chat LLM controllers, and typed decision models.

Traditional control. Fixed-time cycles through a predetermined signal plan without adapting online to measured queues. MaxPressure (MP) chooses the phase with the largest difference between upstream and downstream queue pressure. For intersection $i$ and candidate phase $a$, the implemented phase score $P_i(a)$ is

$$P_i(a) = \sum_{\ell \in \mathcal{L}_i(a)} \max\left(0, q_\ell^{\text{up}} - q_\ell^{\text{down}}\right), \qquad a_i = \underset{a \in \mathcal{A}_i}{\operatorname{argmax}} P_i(a)$$

Here $\mathcal{L}_i(a)$ denotes the set of served traffic movements for phase $a$, and $q_\ell^{\text{up}}$ and $q_\ell^{\text{down}}$ denote upstream and downstream queue measures for movement $\ell$. A third baseline, Greedy-queue, uses the same phase-oriented state (which we will introduce later) as the typed models and selects the phase serving the greatest total incoming queue. It tests whether the supplied phase-level features already support an effective simple decision rule.

RL-based control. We use PPO **(Schulman et al., 2017)** as the learning baseline. The control problem is formulated as an MDP at the 5 s decision epochs. For a single-intersection scenario, the observation at epoch $k$ is the vector.

$$o_k^i = [onehot\ current\ phase\ (4), \min green\ flag\ (1),\ lane\ densities\ (8),\ lane\ queues\ (8)]$$

giving 21 features for 2way and cologne1. For 2x2, the centralized policy concatenates four such observations (84 features); for cologne3, the three intersections contribute 14, 17, and 21 features (52 total). The action is the green-phase index: $a_k^i \in \{0, \dots, 3\}$ for 2way/cologne1/2x2, and a multi discrete vector with cardinalities 3,4,4 for cologne3. The reward is the negative change in accumulated lane waiting time,

$$r_k^i = \frac{1}{100}\left(\sum_{\ell} W_{\ell,k-1} - \sum_{\ell} W_{\ell,k}\right)$$

where $W_{\ell,k}$ is the accumulated waiting time on lane $\ell$; the multi-intersection reward is the mean over intersections. Episodes last 1,800 s (360 epochs). The 2way and cologne1 policies are trained for 100,000 steps; the 2x2 and cologne3 policies for 50,000 and 30,000 steps, respectively. Single-intersection evaluation applies the same max-green wrapper used by the other controllers; the multi-intersection PPO baseline uses the environment's native phase transitions and does not add the separate wrapper. Further hyperparameters are in the supplementary material.

Chat LLM controllers. Each chat model receives the same traffic state, candidate phases, phase option, and safety constraints as the typed models, rendered as a JSON-oriented prompt. A representative prompt has the form:

**Table 2. An example of prompt of chat LLM**

| |
|---|
| Current phase: phase_1. |
| Candidate phases and option: |
| phase_0: serves E->W; queued vehicles 8; longest wait 22 s; last served 15 s ago. |
| phase_1: serves N->S; queued vehicles 12; longest wait 35 s; current phase. |
| phase_2: serves W->E; queued vehicles 5; longest wait 11 s; last served 40 s ago. |
| phase_3: serves S->N; queued vehicles 2; longest wait 60 s; last served 55 s ago. |
| Constraints: min green 5 s; max green 50 s; yellow 2 s. |
| Which phase should receive green next to minimize total vehicle waiting time? |
| Return only a JSON object: {"choice":"phase_2"}. |

The models are invoked in low/no-reasoning configurations, deliberately matched to the typed decision models so that the comparison favors concise and timely action outputs, while we also provide the reason effort sensitivity experiment in the supplement materials, the results don't change our main conclusions. The prompt objective is the same as the PPO reward and the typed question: reduce total vehicle waiting time, subject to the same phase constraints.

Typed decision controllers. Jev and GPT Decisions receive the same underlying traffic observations through an S1 state description. S1 is the phase-oriented input used in the principal comparisons: it organizes lane-level information by candidate phase and includes the current phase, time since activation, queue indicators, waiting indicators, time since last service, and action constraints. The two state descriptions comparison are summarized below in Table 3. For example, instead of presenting an unordered list of lane queues, the S1 input describes "phase 1 serves movements with queue 12 and waiting 35 s; phase 2 serves movements with queue 5 and waiting 11 s" and requests one of the legal phases. The phase returned by the hosted model is passed through the same signal-constraint wrapper.

**Table 3. Two kinds of input to the typed decision model**

| Description | Example input fragment |
|---|---|
| Raw | lanes: [{"id":"N_0","queue_veh":8,"waiting_time_s":35}, {"id":"E_0","queue_veh":5,"waiting_time_s":11}, ...]; current phase: 1; constraints |
| S1 (phase-centric) | phases: [{"phase":0,"movements":"EW","queued_vehicles":8,"longest_wait_s":22," last_served_s_ago":15 },{"phase":1,"movements":"NS","queued_vehicles":12,"longest_wai t_s":35,"current":true}, ...], constraints |

## 4.4. Evaluation and API efficiency

For each scenario, waiting time and throughput are computed for the three traffic seeds, and the tables show mean and standard deviation where repeated runs are available. We separately measure the paid API cost and call latency of each hosted model. The complete suite generates $360 \times 3 \times (1 + 1 + 4 + 3) = 9{,}720$ phase-decision requests per model across the four scenarios and three seeds. We report scenario-level ranges of median and 50th and 95th - percentile response times. A single-intersection decision requires one call; a 2x2 or cologne3 control epoch requires four or three calls, respectively, made sequentially in our implementation.

## 5. Experimental results

### 5.1. Predicting observed travel-mode choices

Table 4 presents the Swissmetro results under the observed-choice instruction F1. MNL achieves the highest top-1 accuracy, 0.688, and the best tested chat LLM, DeepSeek, achieves 0.635. Jev and GPT Decisions obtain 0.546 and 0.514, respectively, compared with a majority-class baseline of 0.586. For this predictive formulation, fitting a task-specific statistical choice model provides a clear accuracy advantage over both typed services. The trained MNL also has essentially negligible execution latency once its parameters have been estimated.

The class-balanced and share metrics provide additional information. DeepSeek has the highest macro-F1 (0.497), while MNL obtains 0.462. The share MAE of MNL is 0.004, smaller than DeepSeek and Claude at 0.094, whereas Jev and GPT Decisions obtain 0.159 and 0.160. The gap between top-1 and macro-F1 reflects class imbalance. SWISSMETRO accounts for 58.6% of the test choices, so a model that favors the majority class can achieve a high top-1 while performing poorly on TRAIN and CAR. Macro-F1 weights the three classes equally, making this failure visible. The majority baseline illustrates the effect: it reaches 0.586 top-1 but only 0.246 macro-F1. Overall, the typed models are less accurate in this observed-choice setting but respond substantially faster than the tested chat LLMs, suggesting that their main advantage here is inference efficiency rather than predictive precision.

**Table 4. Swissmetro observed-choice prediction (1,791 test records)**

| Method | Top-1 | Macro-F1 | Share MAE | API cost ($) | p50 latency |
|---|---|---|---|---|---|
| Majority class | 0.586 | 0.246 | 0.276 | 0 | <1 ms |
| MNL | **0.688** | 0.462 | **0.004** | **0** | **<1 ms** |
| Jev | 0.546 | 0.400 | 0.159 | 0.221 | ~0.38 s |
| GPT Decisions | 0.514 | 0.415 | 0.160 | 0.351 | ~0.68 s |
| DeepSeek V4.1 Flash | 0.635 | **0.497** | 0.094 | 0.489 | ~1.0 s |
| GLM 5.3 Flash | 0.573 | 0.421 | 0.137 | 0.317 | ~1.3 s |
| GPT-6 Luna (chat) | 0.573 | 0.466 | 0.099 | 0.402 | ~1.44 s |
| Claude Haiku 5.5 | 0.587 | 0.423 | 0.094 | 0.632 | ~1.28 s |

*Note: latencies refer to the corresponding reported API calls. Bold identifies the best reported value for each accuracy metric.*

### 5.2. Sensitivity to representation, framing and option order

The choice data allow the same set of alternatives to be described in several ways. We first examine **representation** while retaining the labels and candidate alternatives. Table 5 reports the values from the five-representation subset of the seven-configuration sweep. The difference is substantial: Jev reaches 0.619 under raw R0 but falls to 0.432 under bucket R3; GPT Decisions reaches 0.427 under R0 but only 0.235 under structured R1 and 0.266 under bucket R3. Thus, even when candidate options refer to the same underlying modes, the description affects the selected option and the resulting prediction accuracy.

**Table 5. Reported Swissmetro accuracy under representation variants.**

| Representation | Jev | GPT Decisions |
|---|---|---|
| R0 — Raw attribute strings | 0.619 | 0.427 |
| R1 — Structured fields | 0.604 | 0.235 |
| R2F1 — Relative comparisons, observed-choice instruction | 0.546 | 0.514 |
| R3 — Coarse attribute buckets | 0.432 | 0.266 |
| R4 — Narrative description | 0.562 | 0.455 |

Next, we examine **question framing** and **option order** under the relative-comparison representation R2. Table 6 compares the past-choice question F1, hypothetical-choice question F2, and normative-choice question F3 with different option orders. Jev's mean agreement with the survey label increases from 0.546 to 0.619 and 0.639, respectively. The framing tabulation for GPT Decisions gives 0.514, 0.624 and 0.639, also showing strong sensitivity to wording. At the best reported R2/F3 configuration, both typed models reach approximately 0.639 agreement with Swissmetro labels, comparable to or numerically above the best chat-LLM result (DeepSeek 0.635) listed in Table 4, though still below MNL (0.688). Option order is another relevant input variable. Under R2F3, rearranging the candidate descriptions produces a range of 2.2 percentage points for Jev and 3.9 points for GPT Decisions. For Jev under R2F1, the reported range is much larger—17.5 percentage points—showing that some input configurations are more position-sensitive than others.

**Table 6. Question-framing and option order sensitivity under relative representation R2**

| Model | Framing under fixed R2 | Original | Reversed | Rotated | Mean | Range (pp) |
|---|---|---|---|---|---|---|
| Jev | F1 — did choose | 0.6348 | 0.5433 | 0.4601 | 0.5461 | 17.5 |
| | F2 — would choose | 0.6516 | 0.5980 | 0.6080 | 0.6192 | 5.4 |
| | F3 — is best for | 0.6516 | 0.6298 | 0.6354 | 0.6389 | 2.2 |
| GPT Decisions | F1 — did choose | 0.5293 | 0.4891 | 0.5232 | 0.5139 | 4.0 |
| | F2 — would choose | 0.5980 | 0.6449 | 0.6304 | 0.6244 | 4.7 |
| | F3 — is best for | 0.6198 | 0.6583 | 0.6399 | 0.6393 | 3.9 |

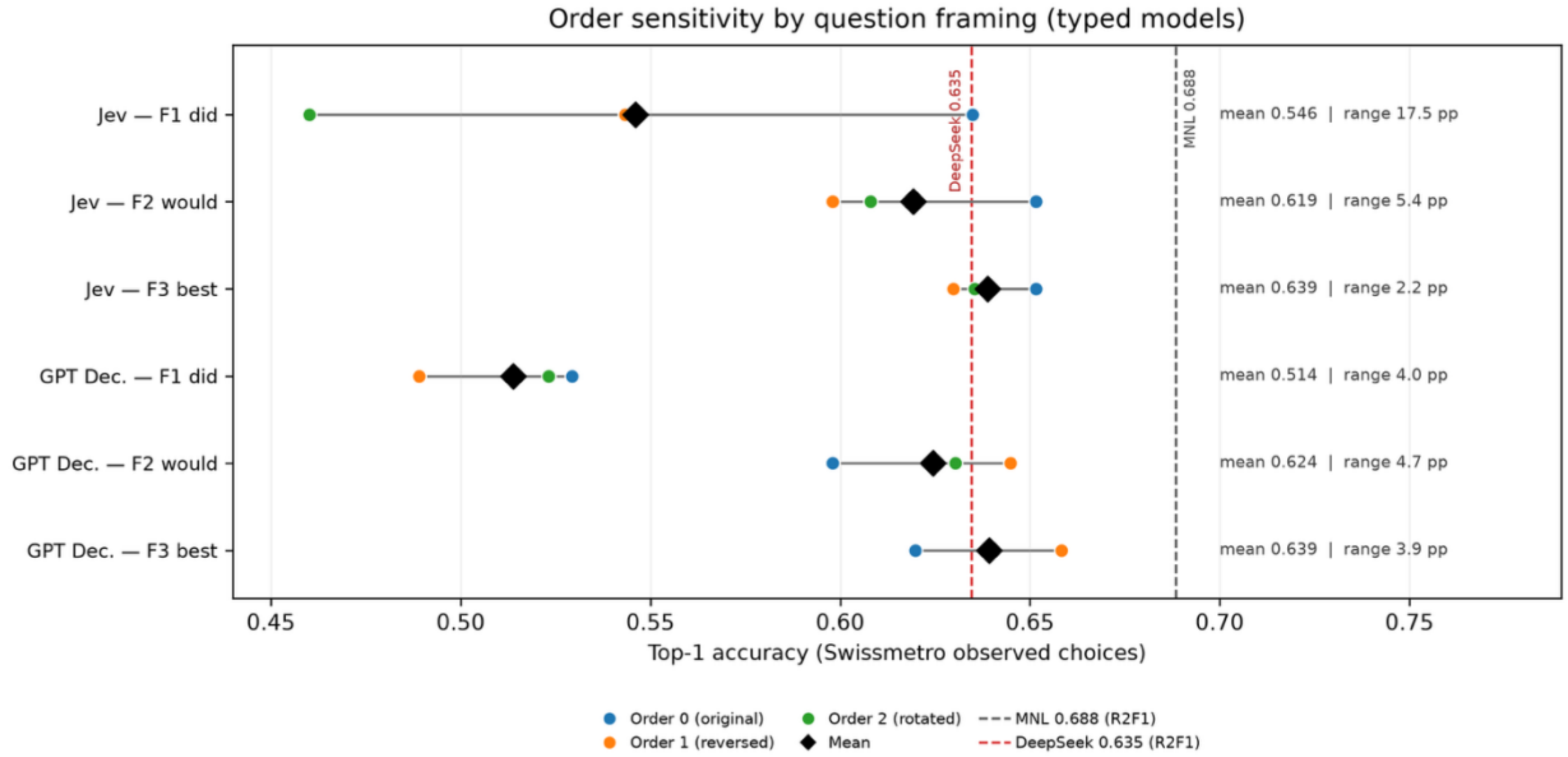

**Figure 2.** Order option sensitivity by question framing for typed decision models. Each row fixes the relative representation R2 and the framing; colored dots are the three option orders, the black diamond is the three-order mean, and the grey bar is the order range. Dashed vertical lines mark MNL (0.688) and DeepSeek (0.635) under the R2F1 observed-choice configuration; they are reference points, not framing-specific results.

Two findings can be concluded. First, typed-model choice accuracy strongly depends on both the representation of alternative attributes and the formulation of the question. A suitable input configuration can bring reported agreement close to the strongest chat LLM, although the observed-choice prediction under F1 remains weaker than MNL. Second, option-order effects are relatively small under the strongest recorded R2F3 configuration but can be substantially larger under another formulation. These results motivate explicit reporting of input configurations whenever decision models are used for transportation prediction.

### 5.3. Closed-loop signal control: waiting time and throughput performance

Table 7 summarizes waiting-time results for the four SUMO environments. Bold identifies the lowest numerical value in each scenario across the traditional, typed and chat controllers; underlining identifies the second-lowest numerical value, including ties. The reported values are means and standard deviations across three traffic seeds. Figure 3 is an illustrative ranking of different methods under various scenarios.

**Table 7a. Mean waiting-time statistic(s): traditional and typed controllers.**

| Scenario | Fixed-time | MP | Greedy-queue | PPO | Jev S1 | GPT Decisions |
|---|---|---|---|---|---|---|
| 2way | 19.40 ± 0.08 | 6.00 ± 0.35 | 6.03 ± 0.25 | 4.16 ± 0.47 | 3.65 ± 0.10 | 4.37 ± 0.33 |
| cologne1 | 8.98 ± 0.04 | 0.48 ± 0.04 | 0.52 ± 0.03 | 0.48 ± 0.03 | 1.21 ± 0.08 | **0.45 ± 0.02** |
| 2x2 | 8.34 ± 0.26 | 0.88 ± 0.07 | 0.36 ± 0.02 | 0.41 ± 0.29 | **0.32 ± 0.01** | 0.73 ± 0.01 |
| cologne3 | 6.24 ± 0.12 | **0.98 ± 0.07** | 53.55 ± 91.51 | 1.64 ± 0.76 | 4.76 ± 4.20 | 62.83 ± 54.01 |

**Table 7b. Mean waiting-time statistic(s): chat LLM controllers.**

| Scenario | DeepSeek | GLM | GPT-6 Luna (chat) | Claude Haiku |
|---|---|---|---|---|
| 2way | **3.32 ± 0.04** | 3.88 ± 0.19 | 3.76 ± 0.06 | 3.40 ± 0.18 |
| cologne1 | 0.71 ± 0.11 | 0.47 ± 0.04 | 0.56 ± 0.05 | 0.60 ± 0.07 |
| 2x2 | 0.77 ± 0.04 | **0.32 ± 0.01** | 0.73 ± 0.05 | 0.55 ± 0.05 |
| cologne3 | 37.26 ± 62.04 | **0.98 ± 0.26** | 1.20 ± 0.45 | 1.25 ± 0.15 |

On 2way, Jev achieves a waiting-time statistic of 3.65 s, outperforming MaxPressure (6.00 s), Greedy-queue (6.03 s), and PPO (4.16 s). It also approaches the best chat LLMs, DeepSeek (3.32 s) and Claude (3.40 s). GPT Decisions (4.37 s) is better than MP and Greedy but slightly behind PPO. These results indicate that typed decisions can supply effective phase selections in a simple single-intersection setting. On cologne1, GPT Decisions records the lowest waiting-time value, 0.45 s, followed by GLM (0.47 s), MP (0.48 s), and PPO (0.48 s). Jev reaches 1.21 s. For 2x2, Jev and GLM obtain 0.32 s, outperforming MP (0.88 s), the same-state Greedy controller (0.36 s), and the available PPO result (0.41 s). GPT Decisions obtains 0.73 s, which is below MP but above Greedy and PPO. Thus, the two typed models show different strengths even under the same phase-oriented input design. The cologne3 results contrast sharply with the smaller networks. MP and GLM both record about 0.98 s; Jev increases to 4.76 s, and GPT Decisions to 62.83 s, with large between-seed variation. Greedy-queue and DeepSeek also exhibit severe outlier runs. In summary, Jev performs strongly on the 2way and 2x2 waiting-time comparisons, and GPT Decisions performs especially well on cologne1. The heterogeneous multi-intersection scenario reveals a deterioration for some typed configurations, suggesting that state information and control robustness become more important in complex networks.

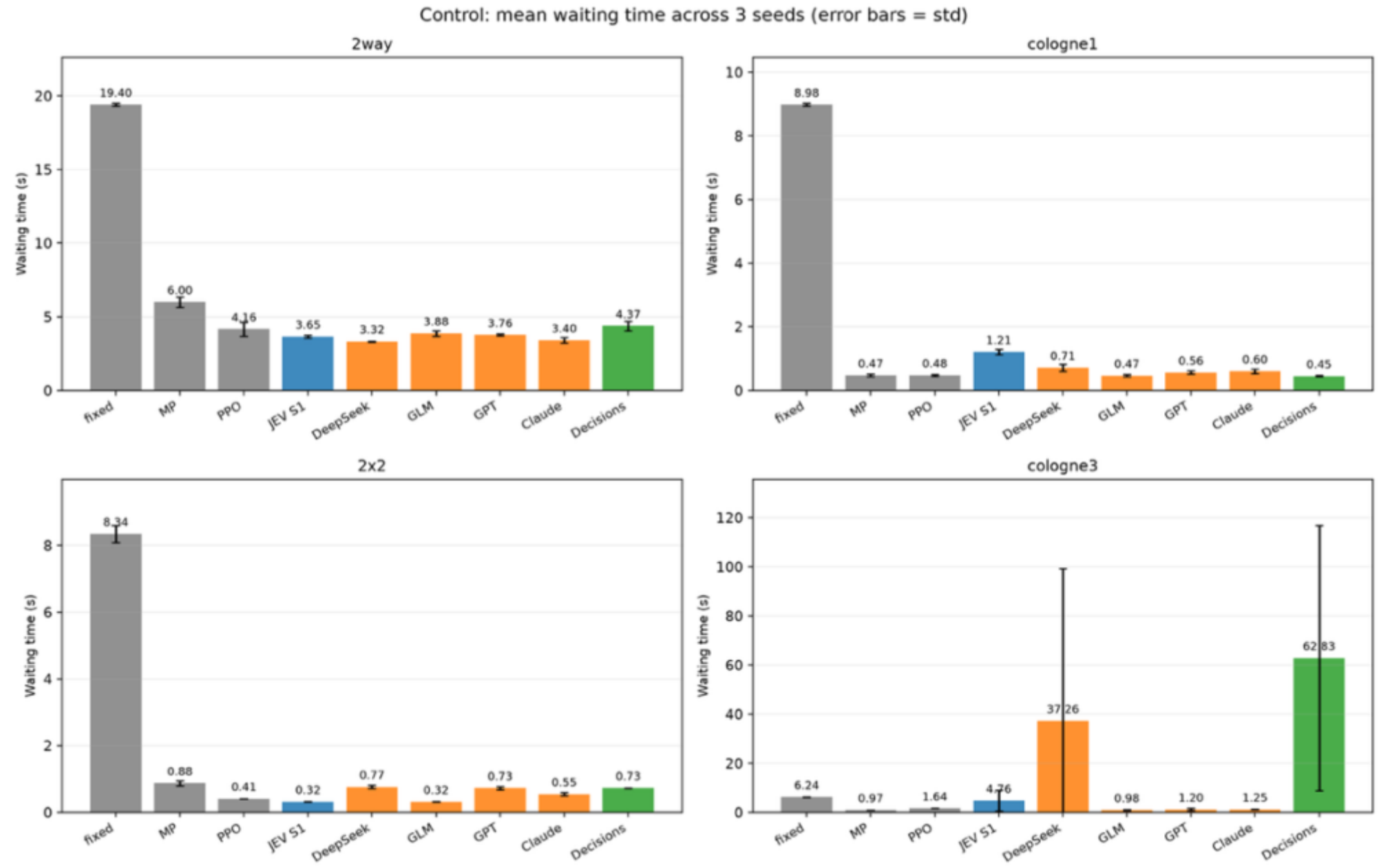

**Figure 3.** Waiting time performance of different control methods

Table 8 reports simulated throughput. GPT Decisions records 413 vehicles/h on 2way, the highest value among the tested methods, and 398 vehicles/h on cologne1, exceeding MP (379), PPO (393), and all four tested chat LLMs, although below fixed-time (438). Its results are lower in the multi-intersection cases, with 228 vehicles/h on 2x2 and 590 on cologne3. Jev's throughput is comparatively stronger in the multi-intersection cases than that of GPT Decisions: it records 244 versus 228 vehicles/h on 2x2, and 720 versus 590 vehicles/h on cologne3. On 2x2, it is also above MP (229), whereas on cologne3 it exceeds both fixed-time (663) and MP (711) numerically, though it remains below the strongest available PPO result (764) and the GPT chat model (733). The throughput results therefore reveal a different pattern from waiting time: GPT Decisions performs well on the isolated intersections, while Jev preserves more completed flow in the more complex networks.

An important advantage of typed decision models over learning-based method like PPO lies in their zero-shot deployment capability. PPO requires scenario-specific policy training, involving 100,000 environment steps for 2way and cologne1 (training time per seed is approximately 35 minutes), 50,000 for 2x2 (training time per seed is approximately 65 minutes), and 30,000 for cologne3 (training time per seed is approximately 25 minutes). In contrast, JEV and GPT Decisions operate without any task-specific parameter training, yet achieve comparable or superior control performance in several scenarios. The ability to make competitive decisions without policy optimization is particularly valuable when adapting to new traffic environments.

**Table 8a. Throughput (vehicles/h): traditional and typed controllers.**

| Scenario | Fixed-time | MP | Greedy-queue | PPO | Jev S1 | GPT Decisions |
|---|---|---|---|---|---|---|
| 2way | 404 ± 17 | 399 ± 17 | 401 ± 13 | 401 ± 13 | 375 ± 12 | **413 ± 38** |
| cologne1 | **438 ± 27** | 379 ± 13 | 347 ± 7 | 393 ± 31 | 370 ± 6 | 398 ± 21 |
| 2x2 | **280 ± 10** | 229 ± 17 | 248 ± 10 | 268 ± 27 | 244 ± 20 | 228 ± 19 |
| cologne3 | 663 ± 9 | 711 ± 44 | 596 ± 36 | **764 ± 14** | 720 ± 40 | 590 ± 89 |

**Table 8b. Throughput (vehicles/h): chat LLM controllers.**

| Scenario | DeepSeek | GLM | GPT-6 Luna (chat) | Claude Haiku |
|---|---|---|---|---|
| 2way | 351 ± 24 | 402 ± 16 | 376 ± 26 | 389 ± 25 |
| cologne1 | 362 ± 17 | 390 ± 32 | 370 ± 16 | 374 ± 11 |
| 2x2 | 235 ± 27 | 261 ± 21 | 245 ± 11 | 258 ± 23 |
| cologne3 | 631 ± 87 | 691 ± 2 | 733 ± 8 | 685 ± 23 |

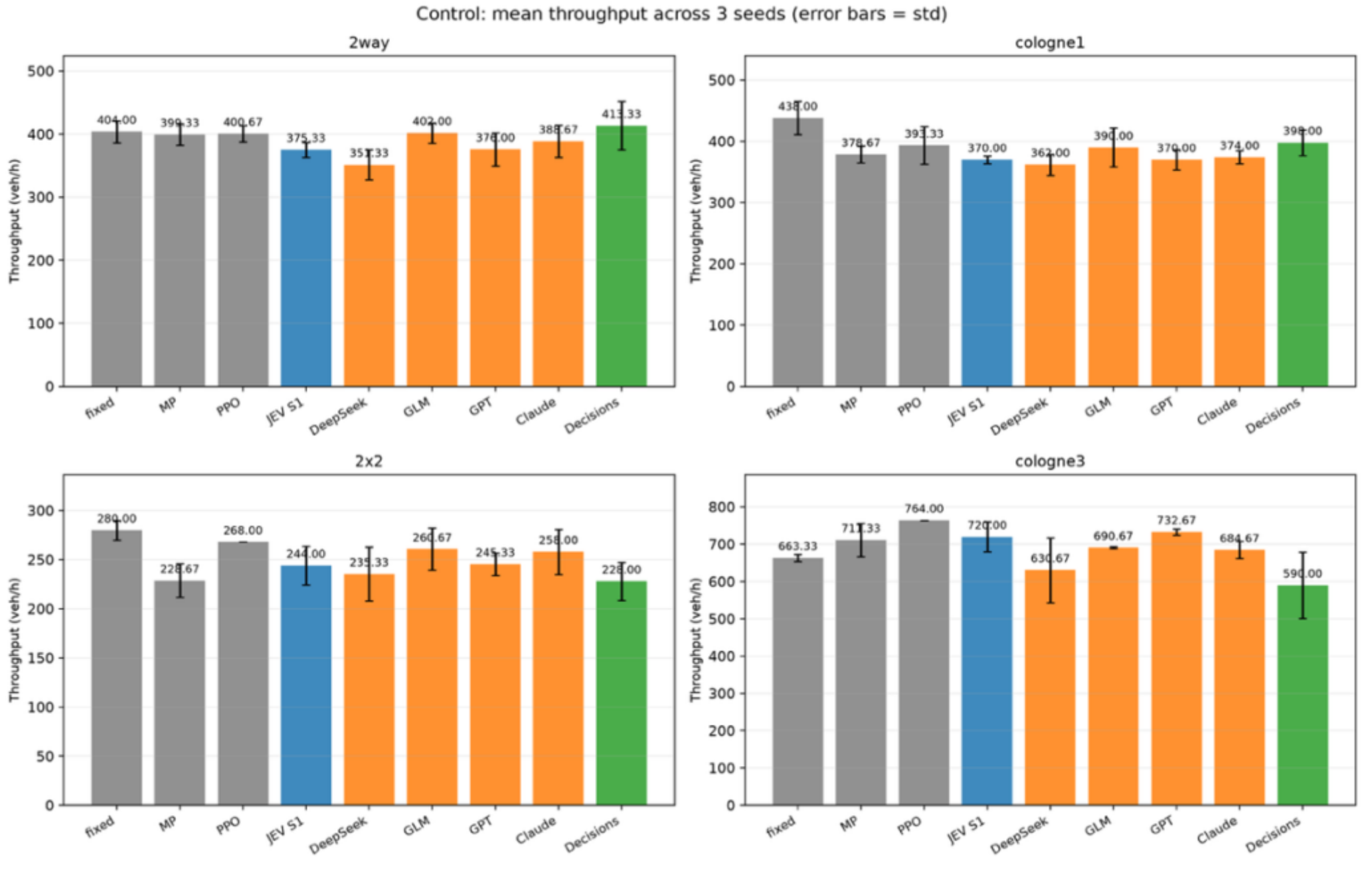

**Figure 4.** Throughput performance of different control methods

### 5.4. Traffic-state design

Table 9 examines two ways of describing the traffic state in 2way. Raw state supplies lane-level observations without reorganizing them around competing green phases. S1 state groups relevant queues and waits by the phases that would serve those movements. It is the standardized phase-oriented representation used for the main model comparisons. The exact fields used in each variant are given in the supplementary input schemas.

**Table 9. State-input sensitivity on 2way: mean waiting-time statistic (s).**

| Model | Raw state | S1state |
|---|---|---|
| Jev | 296.70 ± 24.48 | **3.65 ± 0.10** |
| GPT Decisions | 409.07 ± 72.89 | 4.37 ± 0.32 |

For Jev, the waiting-time value falls from 296.70 s with raw input to 3.65 s with S1 input. GPT Decisions shows a similar improvement, from 409.07 s to 4.37 s. The difference exceeds two orders of magnitude for some comparisons. The main implication is direct: state construction strongly determines the quality of the traffic decisions produced by a typed model. Aggregating and presenting traffic information around admissible signal phases appears particularly useful in the tested intersection. This result also suggests that input-state design should be treated as a substantive part of adapting typed models to a transportation application, rather than a minor formatting choice.

### 5.5. API cost and latency

Table 10 compares the hosted inference services on the control task. Across 9,720 model decisions, Jev costs approximately $0.67 and GPT Decisions $0.98. Jev has the lowest cost of all evaluated hosted models; GPT Decisions costs less than DeepSeek, GPT-6 Luna chat and Claude, while remaining close to GLM ($0.94). The typed services also provide faster median responses than all four chat comparators. For example, Jev's scenario-level median latency ranges from 280 to 425 ms and GPT Decisions' from 611 to 683 ms, compared with 884–1,772 ms across the chat models. A signal-control application must obtain decisions at each intervention step. For an isolated intersection, the controller makes one API call per 5-second simulation decision interval; 2x2 and cologne3 require four and three calls, respectively. In our current implementation these are issued sequentially and then applied together in SUMO. The measured per-call latency establishes an advantage over chat models for individual requests, while total control-cycle latency depends on the number and timing of calls. This is especially relevant to any future real-time implementation. Overall, typed decision models offer a favorable inference speed–cost trade-off relative to most of the tested chat LLMs, with Jev the most efficient hosted service in this comparison. Combined with strong waiting-time and throughput results in selected networks, this supports their potential as lightweight traffic-signal decision modules.

**Table 10. API cost and per-call latency in signal control (9,720 requests per hosted model)**

| Method | Total cost ($) | $/1,000 decisions | p50 per call (ms) | p95 per call (ms) |
|---|---|---|---|---|
| Jev S1 | **~0.67** | **~0.069** | **280–425** | 516–1,608 |
| GPT Decisions | ~0.98 | ~0.101 | 611–683 | **852–990** |
| DeepSeek | ~1.39 | ~0.143 | 884–1,044 | 1,909–2,245 |
| GLM | ~0.94 | ~0.097 | 1,312–1,519 | 2,957–3,554 |
| GPT-6 Luna (chat) | ~1.47 | ~0.151 | 1,421–1,772 | 1,968–2,770 |
| Claude Haiku | ~1.52 | ~0.156 | 1,276–1,592 | 1,628–2,375 |

## 6. Conclusions and future work

This study systematically evaluates typed decision models in transportation choice prediction and dynamic traffic signal control. Three key findings emerge.

First, typed decision models demonstrate promising potential for efficient, training-free transportation decision-making. JEV achieves 0.639 accuracy in travel-choice prediction, comparable to the strongest chat LLM (0.635). Importantly, unlike PPO, which requires tens of thousands of scenario-specific training steps, typed decision models achieve competitive performance in several traffic-control scenarios without task-specific training. The two decision models also provide structured probabilistic outputs with lower inference latency and API costs than most evaluated chat LLMs. This zero-shot capability, together with low inference latency and structured probabilistic outputs, highlights their potential as efficient and adaptable decision-making tools, particularly in settings where retraining is costly or impractical.

Second, input representation is a critical determinant of decision quality. Across different representations of the same travel-choice problem, prediction accuracy varies by 20.7 and 38.5 percentage points for JEV and GPT Decisions, respectively. In traffic signal control, replacing raw traffic states with phase-oriented representations reduces waiting time by approximately 81–100 times. These findings demonstrate that effective representation design can improve model performance without modifying model parameters.

Third, option-order sensitivity is conditioned on question framing. For JEV, changing option order produces a 17.5 percentage points accuracy variation under one question framing, but only 2.2 percentage points under another, despite identical underlying information. GPT Decisions similarly exhibits limited variation (3.9 pp) under the latter framing. These results reveal an interaction between representation and information ordering, indicating that such option order sensitivity can be mitigated through appropriate input design.

The study also identifies three directions for subsequent research. **(i)** Static decision evaluation should be expanded beyond Swissmetro to additional travel choice and transportation prediction tasks with different alternatives and populations. **(ii)** Signal-control benchmarking should include richer network information, more standardized traffic demands and explicit coordination across intersections, particularly for complex multi-intersection networks. **(iii)** Further transportation applications—such as driver behavior, route selection, vehicle maneuver decisions, and other behaviorally grounded finite-action tasks—could examine whether the natural-language understanding of typed models provides additional advantages. These directions would help build a broader benchmark for general-purpose decision models in transportation.

## Acknowledgements

The authors received no financial support for the research, authorship, or publication of this article.

## Declaration of competing interest

The authors declare that they have no known competing financial interests or personal relationships that could have appeared to influence the work reported in this paper.

# Supplementary-material

**Appendix A. API and state examples**

A.1 Prediction task and typed request

JEV request

{ "model": "jev-1.14.0",
  "state": {"context": "A stated-preference survey of intercity travel in Switzerland, 1998. The traveler chose one of the alternatives.",
    "traveler": {"trip_purpose": "business","age_group": "40-54","sex": "male","annual_income": "under CHF 50k","first_class": false,"fare_paid_by": "self","luggage": "none","has_ga_annual_pass": false},
    "units": "Travel time is in minutes; cost is in Swiss francs (CHF); headway is the number of minutes between services."},
  "questions": {"r2_relative_f1": {"type": "choice",
      "instructions": "Which alternative did this traveler choose in this stated-preference survey? Choose the alternative that best matches the traveler's behavior, not the one you would recommend.",
      "option": {"TRAIN": {"travel_time": "213 min","trip_cost": "CHF 100","comparison": "104 min slower and CHF 14 cheaper than SWISSMETRO; 69 min slower and CHF 30 cheaper than CAR","headway": "every 30 min"},
        "SWISSMETRO": {"travel_time": "109 min","trip_cost": "CHF 114","comparison": "104 min faster and CHF 14 more expensive than TRAIN; 35 min faster and CHF 16 cheaper than CAR","headway": "every 10 min", "airline_style_seats": false},
        "CAR": {"travel_time": "144 min","trip_cost": "CHF 130","comparison": "69 min faster and CHF 30 more expensive than TRAIN; 35 min slower and CHF 16 more expensive than SWISSMETRO"}}}}}

JEV response

{"model": "jev-1.14.0",
  "answers": {"r2_relative_f1": {"type": "choice","choice": "SWISSMETRO","provider_choice":
        "SWISSMETRO","choice_mismatch": false,"probabilities": {
        "TRAIN": 0.19,
        "SWISSMETRO": 0.77,
        "CAR": 0.04},"confidence": 0.66,"prob_sum_raw": 1.0},"usage": {"input_tokens": 1004,"output_tokens": 97}}

GPT Decisions uses the same state and questions. Request body is therefore the same as A.1 with that model value.

GPT Decisions response

{"model": "openai/gpt-6-luna-decisions-20261006",
  "answers": {"r2_relative_f1": {"type": "choice","choice": "SWISSMETRO","provider_choice": "SWISSMETRO","choice_mismatch": false,"probabilities": {
        "TRAIN": 0.3,
        "SWISSMETRO": 0.7,
        "CAR": 0.0},"confidence": 0.55,"prob_sum_raw": 1.0},
  "usage": {"input_tokens": 666,"output_tokens": 0,"cost": 6.66e-05}}

A.2 Chat LLM prediction

We use DeepSeek V4.1 Flash as an example. It receives the following system/user messages. The user message is generated from the same state/questions; the chat model returns choices only, without probabilities.

Chat request messages

[{"role": "system",

"content": "You are answering a stated-preference survey. Return only a JSON object mapping every question id to one alternative name. No explanation."},
{"role": "user",
"content": "State:{"context": "A stated-preference survey of intercity travel in Switzerland, 1998. The traveler chose one of the alternatives.", "traveler": {"trip_purpose": "business", "age_group": "40-54", "sex": "male", "annual_income": "under CHF 50k", "first_class": false, "fare_paid_by": "self", "luggage": "none", "has_ga_annual_pass": false}, "units": "Travel time is in minutes; cost is in Swiss francs (CHF); headway is the number of minutes between services."}Questions: Question id: r2_relative_f1nInstructions: Which alternative did this traveler choose in this stated-preference survey? Choose the alternative that best matches the traveler's behavior, not the one you would recommend. Options: {"TRAIN": {"travel_time": "213 min", "trip_cost": "CHF 100", "comparison": "104 min slower and CHF 14 cheaper than SWISSMETRO; 69 min slower and CHF 30 cheaper than CAR", "headway": "every 30 min"}, "SWISSMETRO": {"travel_time": "109 min", "trip_cost": "CHF 114", "comparison": "104 min faster and CHF 14 more expensive than TRAIN; 35 min faster and CHF 16 cheaper than CAR", "headway": "every 10 min", "airline_style_seats": false}, "CAR": {"travel_time": "144 min", "trip_cost": "CHF 130", "comparison": "69 min faster and CHF 30 more expensive than TRAIN; 35 min slower and CHF 16 more expensive than SWISSMETRO"}} Return only a JSON object mapping each question id to the chosen alternative name. Example: {"qid":"TRAIN"}. Do not explain."}]

Chat response

{"model": "deepseek/deepseek-v4.1-flash-1017","raw_response": "{"r2_relative_f1":"SWISSMETRO"}",
"answers": {"r2_relative_f1": "SWISSMETRO"},
"usage": {"input_tokens": 664,"output_tokens": 28,"cost_usd": 6.852e-05}}

A.3 Control task and typed request

The request below uses S1 state, question, and phase option. JEV and GPT Decisions receive the same JSON except for model; the chat model receives the same state/questions/options rendered as messages. All use candidate phases {phase_0, phase_1, phase_2, phase_3}, min green 5 s, max green 50 s, and yellow 2 s.

JEV request (2way, t=120 s, seed 42)

{"model": "jev-1.14.0",
"state": {"scenario": "2way","intersection_id": "t","sim_time_s": 120.0,"current_phase": 0,"current_phase_movements": "N->W, N->S, S->E, S->N","time_in_phase_s": 120,"min_green_elapsed": true,"max_green_exceeded": true,
"lanes": [{"id": "n_t_0","direction": "N","queue_veh": 0,"density": 0.1057,"waiting_time_s": 0.0}, {"id": "n_t_1",……},……],
"constraints": {"min_green_s": 5,"max_green_s": 50,"yellow_s": 2,"decision_interval_s": 5},
"last_action": "phase_0",
"phases": [{"phase": 0,"movements": "N->W, N->S, S->E, S->N","current": true,"queued_vehicles": 0,"longest_wait_s": 0.0,"last_served_s_ago": 0.0},{"phase": 1,……}, {"phase": 2,......},{"phase": 3,……}]},
"questions": {"next_phase": {"type": "choice",
"instructions": "Which green phase should be activated next at this intersection? Choose the phase that minimizes total vehicle waiting time over the episode while respecting the minimum and maximum green constraints.",
"option": {"phase_1": "Phase 1: serves N->E, S->W; queued vehicles on served lanes: 20; longest waiting time: 552s, last served 120s ago.",
"phase_2": "Phase 2: serves E->N, E->W, W->S, W->E; queued vehicles on served lanes: 13; longest waiting time: 419s, last served 120s ago.",
"phase_3": "Phase 3: serves E->S, W->N; queued vehicles on served lanes: 4; longest waiting time: 139s, last served 120s ago."}}}}

JEV response (2way, t=120 s, seed 42)

{"model": "jev-1.14.0",
"answers": {"next_phase": {"type": "choice",
"choice": "phase_1",
"confidence": 0.89,"probabilities": {
"phase_1": 0.93,

"phase_2": 0.06,
"phase_3": 0.01}}},
"usage": {"input_tokens": 1581,"output_tokens": 47}}

GPT decision response (2way, t=120 s, seed 42)

{"model": "openai/gpt-6-luna-decisions-20261006",
"answers": {"next_phase": {"type": "choice",
"choice": "phase_1","probabilities": {
"phase_1": 0.99,
"phase_2": 0.01,
"phase_3": 0},"confidence": 0.99}},
"usage": {"input_tokens": 942,"output_tokens": 0,"cost": 9.42e-05,"is_byok": false,
"cost_details": {"upstream_inference_cost": 9.42e-05,"upstream_inference_prompt_cost": 9.42e-05,
"upstream_inference_completions_cost": 0}}}

A.4 Chat LLM control

We still use DeepSeek V4.1 flash as an example. The chat model receives the same state, question, and options as the typed models, but as a system/user prompt; it returns JSON only.

Chat request messages

[{"role": "system",
"content": "Return only a JSON object with exactly one key "choice" and one of the option IDs as value. Do not explain."}, {
"role": "user",
"content": "State:{"scenario": "2way", "intersection_id": "t", "sim_time_s": 120.0, "current_phase": 0, "current_phase_movements": "N->W, N->S, S->E, S->N", "time_in_phase_s": 120, "min_green_elapsed": true, "max_green_exceeded": true, "lanes": [{"id": "n_t_0", "direction": "N", "queue_veh": 0, "density": 0.1057, "waiting_time_s": 0.0}, {"id": "n_t_1", ……}, {"id": "e_t_1", ……}, ……, "constraints": {"min_green_s": 5, "max_green_s": 50, "yellow_s": 2, "decision_interval_s": 5}, "last_action": "phase_0", "phases": [{"phase": 0, "movements": "N->W, N->S, S->E, S->N", "current": true, "queued_vehicles": 0, "longest_wait_s": 0.0, "last_served_s_ago": 0.0}, {"phase": 1, ……}, {"phase": 2, ……}, {"phase": 3, ……}]} Question: Which green phase should be activated next at this intersection? Choose the phase that minimizes total vehicle waiting time over the episode while respecting the minimum and maximum green constraints. Options:{"phase_1": "Phase 1: serves N->E, S->W; queued vehicles on served lanes: 20; longest waiting time: 552s, last served 120s ago.", "phase_2": "Phase 2: serves E->N, E->W, W->S, W->E; queued vehicles on served lanes: 13; longest waiting time: 419s, last served 120s ago.", "phase_3": "Phase 3: serves E->S, W->N; queued vehicles on served lanes: 4; longest waiting time: 139s, last served 120s ago."}"}]

Chat response

{"model": "deepseek/deepseek-v4.1-flash-1017","text": "{"choice": "phase_1"}","input_tokens": 1201,"output_tokens": 9,"cost_usd": 0.00018465}

A.5 Two typed-model traffic states

The raw state exposes lane-level queues and waits and leaves the lane-to-phase mapping to the model. The S1 state aggregates by candidate phase and adds last-served information.

Raw lane-level state

{"state": {"scenario": "2way","intersection_id": "t","sim_time_s": 120.0,"current_phase": 0,"current_phase_movements": "N->W, N->S, S->E, S->N","time_in_phase_s": 120,"min_green_elapsed": true,"max_green_exceeded": true,"last_action": "phase_0","constraints": {"min_green_s": 5,"max_green_s": 50,"yellow_s": 2,"decision_interval_s": 5},"lanes": [{"id": "n_t_0","direction": "N","queue_veh": 0,"density": 0.1057,"waiting_time_s": 0.0},{"id": "n_t_1",……},……], "downstream_lanes": [{"id": "t_s_1","queue_veh": 0,"density": 0.0},{"id": "t_e_0",……},]},

"question": "Which green phase should be activated next at this intersection? Choose the phase that minimizes total vehicle waiting time over the episode while respecting the minimum and maximum green constraints.",
"option": {
"phase_1": "Phase 1: serves N->E, S->W.",
"phase_2": "Phase 2: serves E->N, E->W, W->S, W->E.",
"phase_3": "Phase 3: serves E->S, W->N."}}

S1 phase-centric state

{"state": {"scenario": "2way","intersection_id": "t","sim_time_s": 120.0,"current_phase": 0,"current_phase_movements": "N->W, N->S, S->E, S->N","time_in_phase_s": 120,"min_green_elapsed": true,"max_green_exceeded": true,"last_action": "phase_0","constraints": {"min_green_s": 5,"max_green_s": 50,"yellow_s": 2,"decision_interval_s": 5},
"phases": [{"phase": 0,"movements": "N->W, N->S, S->E, S->N","current": true,"queued_vehicles": 0,"longest_wait_s": 0.0,"last_served_s_ago": 0.0}, {"phase": 1, ……}, ……],
"lanes": [{"id": "n_t_0","direction": "N","queue_veh": 0,"density": 0.1057,"waiting_time_s": 0.0}, {"id": "n_t_1", ……},],
"downstream_lanes": [{"id": "t_s_1","queue_veh": 0,"density": 0.0}, {"id": "t_e_0", ……}, ……]},
"question": "Which green phase should be activated next at this intersection? Choose the phase that minimizes total vehicle waiting time over the episode while respecting the minimum and maximum green constraints.",
"option": {
"phase_1": "Phase 1: serves N->E, S->W; queued vehicles on served lanes: 20; longest waiting time: 552s, last served 120s ago.",
"phase_2": "Phase 2: serves E->N, E->W, W->S, W->E; queued vehicles on served lanes: 13; longest waiting time: 419s, last served 120s ago.",
"phase_3": "Phase 3: serves E->S, W->N; queued vehicles on served lanes: 4; longest waiting time: 139s, last served 120s ago."}}

**Appendix B. PPO algorithm configuration.**

Single-intersection PPO. We use PPO with an MLP policy, 4 parallel environments, n_steps=512, batch size 256, learning rate $3 \times 10^{-4}$, discount 0.99, GAE 0.95, entropy coefficient 0.01, and 100,000 training steps per seed. Results are from three seeds. Evaluation applies the shared max-green wrapper. Multi-intersection PPO. A single centralized MLP policy observes the concatenated per-intersection observations and outputs a MultiDiscrete phase vector, one phase per intersection. The reward is the mean of the per-intersection rewards. The same PPO hyperparameters are used, with 50,000 steps on 2x2 and 30,000 steps on cologne3. All SUMO rollouts and PPO training were run on a single Apple MacBook Pro with an Apple M1 Pro chip: 8 CPU cores, a 14-core integrated GPU, and 16 GB unified memory, running macOS 26.5.1. The PPO policies are small MLPs and runs on CPU.

**Appendix C. Reasoning effort sensitivity experiment of chat LLM.**

This appendix reports a separate, exploratory experiment on the chat-LLM baselines. The prompt, state, option, and output format were held fixed; only the provider's reasoning-effort setting was varied among low, medium, and high. For C1 we use the R2F1 prompt on a fixed 500-row stratified Swissmetro sample. For C2 we use the 2-way scenario for 30 min. In C1, higher effort improves accuracy modestly: Claude reaches 0.674 at high effort, DeepSeek 0.661 at medium/high, GLM 0.604 at high, and GPT-6 Luna 0.630 at medium. In C2, low effort is best for all four models on waiting time; higher effort does not produce a consistent gain and generally increases cost and latency, as shown in C3. The result supports a conclusion: in these configurations, additional reasoning effort is not a reliable substitute for better state or policy design.

C1. Prediction task, 500-row sample from Swissmetro dataset

| Model | Low | Medium | High | Best |
|---|---|---|---|---|
| Claude Haiku 5.5 | 0.642 | 0.658 | **0.674** | High |
| DeepSeek V4.1 Flash | 0.639 | **0.661** | **0.661** | Medium/High |
| GLM 5.3 Flash | 0.572 | 0.562 | **0.604** | High |
| GPT-6 Luna | 0.606 | **0.630** | 0.628 | Medium |

C2. Control task, 2way scenario

| Model | Low | Medium | High | Best |
| --- | --- | --- | --- | --- |
| Claude Haiku 5.5 | **3.757** | 4.469 | 4.120 | Low |
| DeepSeek V4.1 Flash | **3.989** | 4.273 | 4.032 | Low |
| GLM 5.3 Flash | **3.586** | 3.903 | 4.015 | Low |
| GPT-6 Luna | **3.798** | 3.922 | 3.831 | Low |

C3. Throughput / cost / latency of C2 experiment

| Model | Throughput L/M/H (veh/h) | Cost L/M/H (USD, one seed) | p50 L/M/H (ms) |
| --- | --- | --- | --- |
| Claude | 438 / 424 / 478 | 0.0745 / 0.0901 / 0.1024 | 2307 / 3197 / 3575 |
| DeepSeek | 424 / 402 / 422 | 0.1506 / 0.1545 / 0.1875 | 3330 / 3033 / 3364 |
| GLM | 428 / 412 / 408 | 0.0360 / 0.0375 / 0.0471 | 1466 / 1303 / 1618 |
| GPT-6 Luna | 414 / 422 / 416 | 0.0766 / 0.0817 / 0.0891 | 3251 / 3270 / 4290 |